\documentclass[UTF8]{article}

\title{A Review of 3D Face Reconstruction From a Single Image}
\author{Hanxin Wang \\
University of Electronic Science and Technology of China
}
\date{}
\usepackage{multicol}
\usepackage{indentfirst}
\usepackage{graphicx}

\begin{document}
\maketitle
\setlength{\parindent}{2em}

\begin{abstract}
    3D face reconstruction is a challenging problem but also an important task 
    in the field of computer vision and graphics. Recently, many researchers put 
    attention to the problem and a large number of articles have been published. 
    Single image reconstruction is one of the branches of 3D face reconstruction, 
    which has a lot of applications in our life. 
    This paper is a review of the recent literature on 3D face reconstruction 
    from a single image. 
\end{abstract}

\section{Introduction}
In people's daily life, face is not only a physiological
feature of human beings, but also contains abundant information. 2D face only
has a single pose and illumination, where people may get ambiguous information. 
While 3D face can have a more complete understanding of the face and it contains a large number of
semantic information. 3D face shape is widely used in many areas, such as 
medical care, medical cosmetology, entertainment and security. And it contributes
to other fields such as face alignment, face recognition and face editing. Hence, 3D face
reconstruction is an important task in the field of coumputer vision and graphics.\par

3D face reconstruction now has two main directions. One is single-image 
reconstruction and the other is multi-image reconstruction \cite{wu2019mvf,ramon2019multi,roth2016adaptivemulti,kemelmacher2011face}. 
Because of the limits of 2D structure, single-image reconstruction is more challenging than multi-image 
reconstruction. And single-image reconstruction has more applications in our
life. Therefore, more and more researchers pay attention to this domain and various
approaches have been proposed to tackle this problem. \par

With the development of deep learning, many excellent convolution networks are proposed, 
such as ResNet \cite{he2016deep}, DenseNet \cite{huang2017densely}, U-net \cite{ronneberger2015u}, Facenet \cite{schroff2015facenet}, 
Inception-v4 \cite{szegedy2017inception}, GoogLeNet \cite{szegedy2015going}, VGG \cite{simonyan2014very}, 
Sphereface \cite{liu2017sphereface}. 
In recent years, deep learing is widely used in various fields of computer vision and image processing, including 
image segmentation \cite{yang2021task,xu2020new,yang2020mono,meng2019weakly,shi2018key,yang2020new,meng2017weakly2,yang2020learning,shi2020query}, 
image dehazing/deraining \cite{wei2021non,luo2020single,li2021single,wu2020unified,wei2020single,li2020region,luo2020multi}, 
bject detection \cite{chen2021bal,qiu2020offset,qiu2020hierarchical,qiu2019a2rmnet,chen2020high,li2019simultaneously,li2020codan,li2019headnet} and 
image quality assessment \cite{wu2017generic,wu2020subjective,ma2021remember,wu2017blind,wang2020blind,tang2017deep,huang2016qualitynet,wu2016q}. 
It also has made great contributions to single-image reconstruction.\par

This work is intended as a review of the recent literature on 3D face reconstruction 
from a single image research works. Articles have been choosen among 2016 and 2020, 
in order to provide the most up-to-date view of the single-image 3D face reconstruction. \par

\section{Approaches}
The research on 3D face reconstruction algorithm has been studied by scholars since 
last century. At present, the single-image 3D face reconstruction methods are almost 
based on 3D morphable model (3DMM) and some other methods like shape from shading (SFS), 
UV map, voxel and so on. 
The details of these methods as follows:

\begin{enumerate}
    \item \textbf{3DMM-based methods}:
    3D morphable model was first proposed by Thomas Vetter \textit{et al.} \cite{blanz1999morphable}
    in the article "a deformable model for the synthesis of 3D faces". Up to now, 
    many 3D face reconstruction methods have been developed on the basis of this model. 
    How to get these fitting parameters is the main problem for 3DMM. 
    With the development of deep learning, many methods have been proposed to 
    provide more possibilities for solving the parameter problem.

    \item \textbf{Other methods}:
    Shape from shading is a method for recovering 3D information from a single image proposed 
    by Horn in 1980. SFS uses the change of the normal vector of the smooth object surface, 
    which changes the brightness of the incident light on the object surface and then reflects the shape of the object.
    And recently, some researchers use some image processing methods such as UV map, Epipolar Plane Images to 
    achieve single-image 3D face reconstruction.

\end{enumerate}

\begin{figure}
    \centering
    \includegraphics[width=\textwidth]{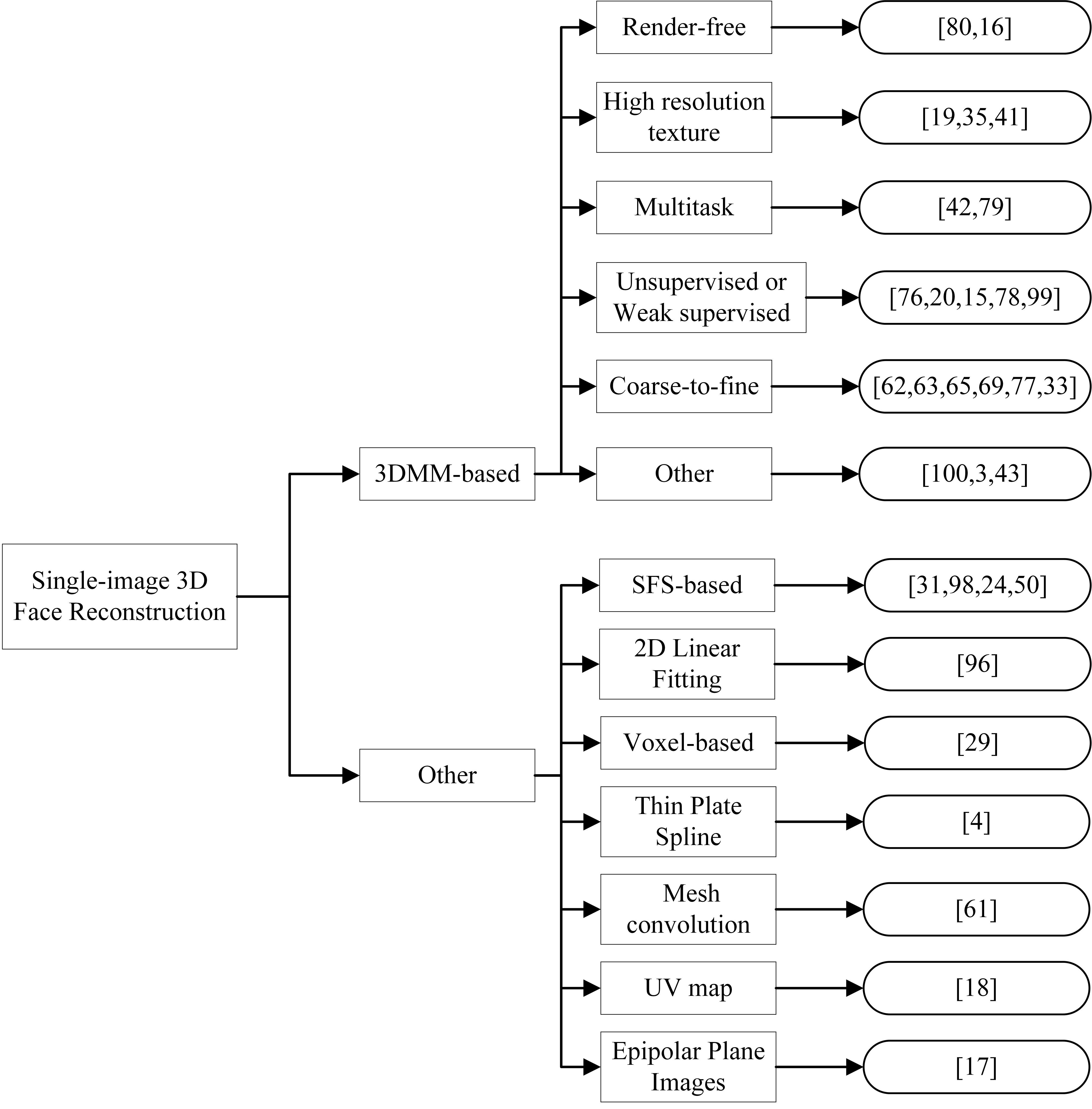}
    \caption{The classification of recent articles}
\end{figure}
At present
\subsection{3DMM-based methods}
3DMM is a statistical model of 3D facial shape and texture. 
It has been widely used in 3D face reconstruction and face recognition \cite{blanz2003face,amberg2008expression,hu2016face,pay20093d}. 
The main contribution of morphable model is to build a correspondence between the reconstruction and other models.
The traditional 3DMM uses mean shape and linear combination of a set of shape bases to 
generate a personalized 3D face shape. The shape bases are typically extracted from a training set of 
3D face scans by Principal Component Analysis (PCA). Through searching for the optimal linear fitting parameters, a reconstructed 
face rendering from 3D face will be approximate to original image. \par

Modeling highly variable 3D face shapes requires a quantity of high-quality 3D face scans.
The first 3DMM was built from scans of 200 subjects with a similar ethnicity/age group. The widely used
Basel Face Model (BFM) \cite{paysan20093d} is an extension of traditional 3DMM. It also built with only 200 
subjects but add some expressions parameters. The new version BFM \cite{gerig2018morphable} is optimized on the expression base.
Later more 3DMMs \cite{booth20163d,ploumpis2020towards,ploumpis2019combining,booth20173d} have been built. 
Among these 3DMMs, \cite{booth20163d} build the first large-scale 3DMM from scans of 10,000 subject 
which can meet the needs of completely describing human face. \par

Because most algorithms are designed for faces in small to medium poses, lacking the ability to align faces in large
poses up to 90°. In 2016, Zhu \textit{et al.} \cite{zhu2016face} propose an new alignment framework, called
3D Dense Face Alignment (3DDFA), to solve the problem. The author builds a cascaded-CNN to predict
3DMM parameters with a specifically designed feature, called Projected Normalized Coordinate Code (PNCC).
Given an initial parameter $ p^k $, the projected normalized coordinate code was generated according to this $ p^k $.
And then stack the projected normalized coordinate code with the input image 
and send it into CNN to predict the parameter $ \Delta p^k $ which is used to update $ p^k $. 
This method achieves significant improvement in large pose.
However, the generated 3D face contains less details.\par

Richardson \textit{et al.} \cite{richardson20163d} employed an iterative CNN trained with synthetic data to estimate 3DMM
parameters. The predicted geometry was then refined by the real-time shape-from-shading method.
In order to extract more details, they extend the work and propose an end-to-end CNN framework which has a coarse-to-fine structure \cite{richardson2017learning}. 
The proposed architecture consists of two main blocks, a CoarseNet and a FineNet. 
Given a image, CoarseNet will output its coarse facial geometry by fitting a 3DMM model. And then CoarseNet connects
with a novel layer which renders a depth image from a 3D mesh. FineNet receives a coarse depth map and stack it with the original
input images. A shape-from-shading method is applied as a refinement to capture the fine facial details. This method allows the network 
to extract more details when a high quality input image is available. And it's robust to expressions and different
poses. But it might fail when tested upon unique facial features such as beards, makeup, and glasses. And some details 
like wrinkles can't be reconstruct. \par

Roth \textit{et al.} \cite{roth2016adaptive} also propose a coarse-to-fine method to archieve 3D face reconstruction with albedo information. 
They utilize a 3DMM to get a coarse 3D face template and then develop a novel photometric stereo formulation to estimate normals in 
the face surface. A personalized 3D face with details can be reconstructed by the normals. But the 3D faces reconstructed by the method 
are not smooth and have low resolution.\par

In 2017, Bas \textit{et al.} \cite{bas20173d} show how a spatial transformer network \cite{jaderberg2015spatial} can be used to generate 3D face with 3DMM. 
The author use the localiser to predict 3DMM shape 
parameters and pose. According to the predicted parameter, the grid generator projects the 3D geometry to 2D. 
At the same time, an occlusion mask is computed from the estimated 3D geometry.
And then a bilinear sampler resamples the input image to a regular 
output grid which is finally masked by the previously obtained occlusion mask. 
This method is able to perform robust on large pose changes images. But it has the same problem as \cite{richardson2017learning}.\par

While Tr$\widetilde a$n \textit{et al.} \cite{tuan2017regressing} focus on the problem when applied “in the wild”, existing methods are either 
unstable or change for different photos of the same subject. To handle this problem, they 
use a CNN to regress 3DMM shape and texture parameters directly. This CNN is a render-free 3DMM estimator and do not 
need to train. The author uses large quantities of unconstrained photos to fit a single 3DMM for each subject. 
Then, all 3DMM estimates for the same subject are pooled together for a single estimate. These pooled estimates are
used to train a very deep CNN to regress 3DMM shape and texture parameters directly. Hence, this method 
is fast and robust to produce similar discriminative 3D shapes for different views. The experimental results show 
the areas around the nose and mouth in particular have very low errors compared with 3DDFA.\par

And Dou \textit{et al.} \cite{dou2017end} also propose a render-free approach. They find the neutral 3D facial shape favors 
higher layer features. While the expressive 3D facial shape favors lower or intermediate layer features. According to the characteristic,
they propose a DNN-based approach for End-to-End 3D FAce Reconstruction (UH-E2FAR) from a single 2D image, which divides 3D face reconstruction 
into neutral 3D facial shape reconstruction and expressive 3D facial shape reconstruction. UH-E2FAR is based on the VGG-Face model \cite{parkhi2015deep}. 3D shape 
parameters are predicted by the VGG-Face model directly. Specially, it adds a sub convolutional neural network that concatenates features from 
intermediate layers of VGG-Face to predict the expression parameters. The performance of our method is better than before and 
the reconstructed expression is more plausible.\par

Sela \textit{et al.} \cite{sela2017unrestricted} propose an Image-to-Image translation network that can translate the input image 
to a depth image and a facial correspondence map. They utilize the two maps to warp a template mesh in the three-dimensional 
space obtained by 3DMM through an iterative non-rigid deformation procedure. And finally, a fine detail reconstruction algorithm 
with the input image recovers the subtle details of the face. The 3D face reconstructed by this method contains 
more details so that look more realistic and is robust to expressions.\par

Training deep neural networks usually requires a great quantity of datas, but face images with 3D ground truth shapes are hardly available.
An autoencoder network was proposed by Tewari \textit{et al.} \cite{tewari2017mofa}, which can be trained on unlabeled photographs to predict shape, 
expression, texture, pose, and lighting simultaneously. The encoder is a regression network to predicte 3DMM parameters, 
and the decoder is a differentiable render layer to reproduce the input photograph. This approach does not require supervised training pairs. 
However, since the training loss is based on individual image pixels, the network is vulnerable to confounding variation between related variables.

Genova \textit{et al.} \cite{genova2018unsupervised} also propose an unsupervised training for 3D face reconstruction in 2018. Similar to \cite{tewari2017mofa}, they 
employ an encoder-decoder architecture. Specially, it exploits a pretrained face recognition network, which distinguishes such related variables 
by extracting and comparing features across the entire image. First, the model is trained on batches of synthetic faces generated by
randomly sampling for shape, texture, pose, and illumination parameters. Second, the partially-trained model is trained to convergence on 
batches consisting of a combination of real face images from the VGG-Face dataset and synthetic faces. When training the partially-trained model, 
the author use three novel losses: a batch distribution loss, a loopback loss and a multi-view identity loss.
The batch distribution loss encourages to match the distribution between the output and the morphable model, 
The loopback loss ensures the network can correctly reinterpret its own output. While the multi-view identity loss try to keep the consistency of 
the predicted 3D face features and the input photograph features from multiple viewing angles. \par

This approach improves on the likenesses of previous approaches, especially in features relevant to facial recognition such as the eyebrow texture and
nose shape. Additionally, the network can reconstruct 3D face from non-photorealistic artwork, in cases where a fitting approach based on inverse rendering
would have difficulty. Compared to MoFA, this approach is more resistant to confounding variables such as identity, expression, skin tone and lighting.\par

In 2019, Deng \textit{et al.} \cite{deng2019accurate} propose a novel deep 3D face reconstruction approach that utilize 
low-level and perception-level information for weakly-supervised learning. Given a training RGB image , they 
use R-Net \cite{2017R} to regress 3DMM parameters. With the parameters, a reconstructed image can be analytically generated by some simple, differentiable math derivation.
This R-Net is trained with evaluating a hybrid-level loss on the reconstructed image and original image instead of ground truth labels.
The hybrid-level loss consists of image-level loss, which contains 2D landmarks loss and skin-aware photometric loss, and perception-level loss, which is 
to extract the deep features of the images and compute the cosine distance by a pre-trained deep face recognition network. The results obtained with 
this method can be significantly better than those trained with synthetic data or pseudo-ground truth shapes. \par

This method is fast, accurate, and robust to occlusion and large pose. Compared with \cite{zhu2016face, genova2018unsupervised,tuan2017regressing}, 
the texture and shape exhibit larger variance and are more consistent with the inputs. And it performs better than PRN \cite{feng2018joint} 
at all views.\par

Tran \textit{et al.} \cite{tran2019learning} consider the linear bases limits the representation power of 3DMM. Therefore they 
propose an innovative framework to learn a nonlinear 3DMM model from a large set of in-the-wild face images.  
The entire network is end-to-end trainable with only weak supervision.
It contains one encoder and two decoders, which serve as the nonlinear 3DMM. Given a face image as input, the encoder estimates 
the projection, lighting, shape and albedo parameters. Each decoder takes a shape or albedo parameter as input and output 
the dense 3D face mesh or a face skin reflectant. They design a differentiable rendering layer to generate a reconstructed face 
by fusing the 3D face, albedo, lighting, and the camera projection parameters. Finally, by minimizing the difference between 
the reconstructed face and the input face, an accurate 3D face can be obtained. This method can reconstruct with smaller errors than 
the linear model and better reconstruct the facial texture. It can faithfully resemble the input expression and significantly 
surpass PRN and 3DDFA.\par

Motivated by \cite{tran2019learning,ranjan2018generating} , Zhou \textit{et al.} \cite{zhou2019dense} propose a non-linear 3DMM by joint learning texture autoencoder 
and shape autoencoder using direct mesh convolutions. A 2D convolution network is used to encode the 
in-the-wild images to obtain joint texture and shape features followed by a mesh decoder. And then use mesh convolutions to generate 
texture and shape. The model is very light-weight and perform Coloured Mesh Decoding (CMD) in-the-wild at a speed of over 2500 FPS.\par

Hassner \textit{et al.} \cite{tran2018extreme} describe a system designed to provide detailed 3D reconstructions of faces  
under extreme conditions like occlusions. Motivated by the concept of bump mapping \cite{blinn1978simulation} , which can 
separate global shape from local details, they propose a layered approach which decouples estimation of a global shape 
from its mid-level details. First they use deep 3DMM approach \cite{tuan2017regressing} 
to estimate a coarse 3D face shape as a foundation and then separately add details represented by a bump map on this foundation. 
Specifically, they use an face segmentation method to determine occluded regions in the input image 
and search for one suitab similar individual in the reference set. Then replace the occluded regions with the details transfered 
from the bump map associated with the selected reference image. This method produces detailed 3D face shapes 
in viewing conditions where existing state of the art often break down.\par

Inspired by works in face landmark marching, Liu \textit{et al.} \cite{liu2019single} develop a method for 3D face reconstruction 
using a novel landmark updating optimization strategy. They get 3DMM parameters by updating contour landmarks and self-occluded 
landmarks instead of predicting directly. For contour landmarks, the detected landmarks on 2D image are to updated by minimizing 
the correspondence error between landmarks projected by 3D landmark points and original 2D landmarks. For self-occluded landmarks, 
they render the model into image plane, extract the edge of projected area, and generate new correspondence landmarks according 
to the edge pixels. This method has lower the reconstruction error than 3DDFA. But they look similar on qualitative results.\par

Gecer \textit{et al.} \cite{gecer2019ganfit} propose a GANFIT method that can reconstruct high-quality texture and shape
from single “in-the-wild” images. GANFIT can be described as an extension of the original 3DMM fitting strategy. But 
instead of a PCA texture model, it uses a Generative Adversarial Network (GAN) to obtain a high-resolution 
texture map. And it uses a state-of-the-art face recognition network \cite{deng2019arcface} to detect face identity in the reconstructed face. 
This method has excellent results in photorealistic 3D face reconstructions and achieve 
facial texture reconstruction with high-frequency details for the first time. \par

Bulit upon GANFIT method, Lattas \textit{et al.} \cite{lattas2020avatarme} propose AvatarMe, a method that is able to 
reconstruct photorealistic 3D faces from a single “in-the-wild” image and contains an increasing level of details. 
A 3DMM is first used to reconstruct a 3D face that only has basic texture from a single image at a low resolution and 
a completed UV texture is synthesized. The basic texture map, which contains baked illumination, 
is enhanced by a super resolution network. Then a de-lighting network obtain a high resolution diffuse
albedo from the high solution texture. Finally, they infer specular albedo, diffuse normals and specular normals from the diffuse albedo in conjunction
with the base geometry. Hence, AvatarMe can keep consistent on varying lighting conditions and produce more photorealistic 3D faces.\par

GAN used in above methods needs a lot of datas to train, but face texture datas is hardly available. Therefore, Lin \textit{et al.} \cite{lin2020towards} 
propose a coarse-to-fine method to reconstruct 3D facial shapes with high-fidelity textures from single-view images in-the-wild, 
without the need to capture a large-scale face texture database. They utilize a regressor regresses the 3DMM coefficients, 
face pose and lighting parameters from the input image. And then they use a PCA model to compute the face shape
and coarse texture from the 3DMM coefficients. Meanwhile the FaceNet is used to extract a face embedding from the input image. Then the coarse texture 
is fed into the GCN Refiner and the face embedding is fed into GCN Decoder. The outputs of the two GCNs are concatenated 
fed to the Combine Net. Finally a fine texture will be produced in the Combine Net. Compared with GANFIT, the results are more photometric. \par

In 2021, Khan \textit{et al.} \cite{khan2021learning} propose two cascaded CNNs in a coarse-to-fine strategy for actual detailed 3D face 
reconstruction from a single image. The cascaded CNNs are composed of two network, a coarse layer network, that generates a coarse-scale shape 
by fitting a 3DMM model and estimates pose and projection parameters for projecting the face shape into the
image plane, and a fine layer network , that get a detailed geometry by replacing every pixel along the depth direction in the face region. 
This method is efficient to pose, expression and lighting dynamics and enhances the ability of capturing facial details like wrinkles.\par

And recently some works propose multitask approach, which accomplish the task of reconstructing accurate 3D face shapes and other task
simultaneously. Liu \textit{et al.} \cite{liu2018disentangling} propose an encoder-decoder network to accomplish 3D face reconstruction and face recognition at the same time. 
The encoder network converts the input 2D face image to identity representations and residual latent representations. 
While the decoder network recovers its 3D face shape from these two representations and mean shape. Meanwhile, identity representations can be 
used for face recognition. The method expands the capacity of 3DMM for capturing discriminative shape features, and it improves accuracy 
both in 3D face reconstruction and in face recognition. It performs better than 3DDFA under different expressions and different 
yaw angles. However, the authors find that optimaling for reconstruction might limit the discriminativeness of shape parameters.
Hence, the method chooses to ensure the superior performance in face recognition.\par

While Tu \textit{et al.} \cite{tu20203d} develop a method that accomplish 3D face reconstruction and dense face alignment simultaneously.
They propose a novel 2D-assisted self-supervised learning (2DASL) method that can effectively improve 3D face model learning when applying in “in-the-wild” face images. 
2DASL model contains 3 modules: a CNN-based regressor, an encoder and a self-critic.
The CNN-based regressor is used to predicts 3DMM coefficients from the input 2D image. The encoder converts the input image 
into a latent representation. While self-critic is designed to determine whether latent representation and 3DMM coefficients pairs 
are consistent or not. The model takes as input the face images with 3D annotations and other images with only 2D Facial Landmark Map (FLM). 
And then the CNN-based regressor predicts two sets of 3DMM coefficients separately. One with 3D annotations is for 3DMM for 3D 
reconstruction and dense alignment and is trained through 3D annotation supervision. The other one only with 2D FLM is trained 
through self-critic supervision. This method produces 3D reconstruction and dense face alignment results with strong robustness 
to large poses and occlusions. \par

\subsection{Other methods}
\textbf{Shape From Shading Based} \quad
SFS used to be a popular approach for 3D face reconstruction from a single image. This method reconstructs 3D face by making assumptions 
on the different light sources and the reflectance properties of face, where it is only available under certain conditions, 
namely Lambertian reflectance and uniform albedo. \par

Kemelmacher \textit{et al.} \cite{kemelmacher20103d} build a reference model under the guidance of input picture to 
align with the face image and then use SFS method to add more details to the reference model. 
Zhao \textit{et al.} \cite{zhao2012human} propose a fast-3D-reconstruction method based on image processing and SFS.
Wavelet transform is used in image denoising first. And then they use SFS to reconstruct 3D face.
Han \textit{et al.} \cite{han2013high} estimated lighting variations with global and local light models. Then utilize
SFS approach and the estimated lighting models to reconstruct accurate shape. 
ROr-El \textit{et al.} \cite{or2015rgbd} proposed an improved SFS method that can create more details. 
They use a new depth map in the SFS by combining the RGB image and rough depth image. \par

SFS is a traditional approach for 3D face reconstruction. The advantage of this approach is less time consuming, 
a good ability of capturing details and only needing a single image .
However, it has a lot of complex math formulas and conditions constrains, which limits the development.\par

\textbf{2D Linear Fitting Based} \quad
Inspire by 3DMM, Yun \textit{et al.} \cite{yun2017cost}
propose a approach based on the idea that a particular coefficient that linearly fits 2D face images can be used to fit 3D models.
They try to Find a set of coefficients that can be linearly combined with 2D images 
so that multiple 2D images approximately synthesize the input target image, 
and then use the coefficients in the 3D model to synthesize the 3D model of the target image.
This approach reduces the complexity by reusing the coefficients obtained after morphing in the 2D image space. And the results show successful
reconstruction of details of the 2D query images, such as eyebrows, lip, skin color, and eye shape. However, because of the 
limit of dataset, skin troubles like freckles, acne, and so on are not entirely recovered and the reconstructed shapes look unnatural.\par

\textbf{Voxel Based} \quad
In order to address the limitations of the requirement of complex and inefficient pipelines for model building and fitting, 
Jackson \textit{et al.} \cite{jackson2017large} propose a CNN-based Volumetric Regression Network - Guided (VRN - Guided). 
The network has an encoder-decoder structure where a set of convolutional layers are firstly used to detect 
the 2D projection of the 3D landmarks and stacks these with the original image. Then this stack is fed into 
the reconstruction network, which directly regresses the volume. This method is simple and can work with totally unconstrained 
images downloaded from the web, including facial images of arbitrary poses, facial expressions and occlusions. But it can not 
capture fine details and the reconstructed 3D faces look coarse so that the accuracy of VRN - Guided still need to improve.\par

\textbf{Thin Plate Spline Based} \quad
Bhagavatula \textit{et al.} \cite{bhagavatula2017faster} propose a novel approach to reconstruct the 3D shape of faces by learning 
a 3D Thin Plate Spline (TPS) warping function. Given a image, a feature extraction network is used to capture face feature first. 
And then the TPS Localization Network estimates TPS parameters to build TPS warping function which warps a generic 3D model 
to a subject specific 3D shape. \par

\textbf{Mesh Convolution Based} \quad
Ranjan \textit{et al.} \cite{ranjan2018generating} consider traditional models like 3DMM, who learn a latent representation of a
face using linear subspaces, can not capture extreme deformations and non-linear expressions. 
To address this, they propose a Convolutional Mesh Autoencoder (CoMA) with mesh sampling operations that 
enable a hierarchical mesh representation. By utilizing mesh sampling operations, non-linear variations can be captured in shape and expression.\par

\textbf{UV map Based} \quad
Feng \textit{et al.} \cite{feng2018joint} propose a straightforward method called Position map Regression Network (PRN)
that can archieve 3D facial structure reconstruction and dense face alignment simultaneously. 
They design a 2D representation called UV position map. The position map records the corresponding spatial position 
of each pixel on the texture map, which can be understood as representing the coordinate \textit{x, y, z} with three channels of \textit{r, g, b}. 
And then they train an encoder-decoder to regress it from a single 2D image. Compared with VRN - Guided,
the method contains more details. Meanwhile, the network is very light-weighted and spends only 9.8ms to process an image, 
which is extremely faster than previous works.\par

\textbf{Epipolar Plane Images Based} \quad
Gilani \textit{et al.} \cite{feng20183d} propose to exploit the Epipolar Plane Images (EPI) \cite{goodfellow2014generative} obtained from light field cameras 
to reconstruct 3D face. They use horizontal and vertical EPIs to train two FaceLFnets separately which can  
output horizontal and vertical 3D facial curves. Then they merge horizontal and vertical 3D facial curves 
into a single pointcloud based on the camera parameters and recover the final 3D face by using a surface fitting method. 
It is a model-free approach and it can estimates the peripheral regions of the face such as hair and neck. 
It reduces reconstruction errors by over 20$\%$ compared to 3DDFA and VRN - Guided.\par

\section{Performance Analysis}

SFS as the traditional single-image 3D reconstruction approach is fast and detailed. However, it has some 
conditions constrains which limit its application. At present, most researchers achieve single-image 3D face 
reconstructions upon 3DMM.\par

Among the 3DMM-based methods, \cite{zhu2016face} shows good performance in large pose up to 90° but the reconstructed face is 
not detailed. \cite{richardson2017learning,roth2016adaptive,khan2021learning} are all a coarse-to-fine architecture and use photometric stereo approaches 
to extract more details. But \cite{khan2021learning} enhances the ability of capturing facial details like wrinkles, while \cite{richardson2017learning}can not. 
And the reconstructed results in \cite{roth2016adaptive} are not smooth and have low resolution. \cite{tuan2017regressing,dou2017end} are both render-free approach 
which can reduce a lot of time spending on rendering and they are is robust to expressions. 
\cite{tewari2017mofa,genova2018unsupervised,deng2019accurate,tran2019learning,zhou2019dense} are unsupervised or weakly-supervised methods. Compared to \cite{tewari2017mofa}, 
\cite{genova2018unsupervised} is more resistant to confounding variables such as identity, expression, skin tone and lighting. 
\cite{deng2019accurate} is fast, accurate, and robust to occlusion and large pose and the texture and shape exhibit larger variance 
than \cite{genova2018unsupervised}. \cite{tran2019learning} is a nonlinear 3DMM model. While \cite{zhou2019dense} is very light-weight than the method 
based on DCNN. \cite{tran2018extreme} can produce detailed 3D face shapes in occlusion. \cite{gecer2019ganfit,lattas2020avatarme,lin2020towards} do well in 
extract face texture so that they are all able to reconstruct photorealistic 3D faces. However, \cite{gecer2019ganfit,lattas2020avatarme} based on 
GAN needs a lot of datas which are unavailable to train. And the results in \cite{lattas2020avatarme,lin2020towards} are more photometric. 
\cite{liu2018disentangling,tu20203d} are both multi-task and show the efficiency on 3D face reconstruction task and other face task. \par

And there are some other methods to achieve good 3D face reconstruction. Among other methods, \cite{yun2017cost} is successful in 
reconstruction of details such as eyebrows, lip, skin color, and eye shape. But the limit of databset makes the results unnatural.
\cite{jackson2017large} is based on voxel. its network is simple but the accuracy is not high. 
\cite{feng2018joint} use the UV map to generate 3D face. Compared with \cite{jackson2017large}, it contains more details and and spends only 9.8ms to process an image.
\cite{feng20183d} is a model-free approach and predicts the peripheral regions of the face such as hair and neck. But the input images limit to light field images. \par

\section{3D Face Datasets}
In the section, some common 3D face datasets are introduced which can be used in 3D face reconstruction, 
face recognition, face alignment and so on. \par

\textbf{AFLW2000-3D} \quad This dataset is built by zhu \textit{et al.} \cite{zhu2016face} in the 3DDFA. 
It is formed by the first 2000 images from AFLW \cite{koestinger2011annotated} and theirs 3D annotations.
In addition to fitted 3DMM parameters, 3D annotations also have 68 3D landmarks. 
Therefore it can be used in both face reconstruction and face alignment tasks. \par

\textbf{BU-3DFE} \quad The BU-3DFE dataset \cite{yin20063d} is a three-dimensional 
facial expression database, which includes three-dimensional 
facial expression shapes and theit corresponding two-dimensional facial textures of 2500 models.
The images are collected from 100 subjects who are from different races. 
Each subject has seven expressions.\par

\textbf{BU-4DFE} \quad The BU-4DFE dataset \cite{zhang2013high} is the extension of the BU-3DFE dataset. 
It presents a high-resolution 3D dynamic facial expression database, where 
the 3D facial expressions are captured at a video rate. It consists of 606 
3D facial expression sequences from 101 subjects. Like BU-3DFE, each subject has seven expressions. \par

\textbf{MICC} \quad The MICC dataset \cite{bagdanov2011florence} is a 3D face dataset captured 
from a high solution 3D scanning system. It contains 53 subjects with theirs ground truth 3D mesh.
In addition, it has some video sequences captured at different resolutions, conditions and scale levels. \par

\textbf{Bosphorus} \quad The Bosphorus dataset \cite{savran2008bosphorus} is a dataset 
released in 2009 collected from structured light. This is intend for studying three-dimensional 
facial expressions. It consists of 4666 faces from 105 subjects. 
Each subject has up to 35 expressions and head poses at different angles. \par

\textbf{FaceWarehouse} \quad The FaceWarehouse dataset \cite{cao2013facewarehouse} is also
a 3d facial expression database. Unlike the above data sets, it is built from 150 
Chinese aged from 7 to 80. Each subject contains 47 different expressions.\par

\textbf{4DFAB} \quad The 4DFAB dataset \cite{cheng20174dfab} is a large scale 
database of dynamic high-resolution 3D faces. It contains over 1,800,000 3D meshes 
collecting from 180 subjects in four different sessions. \par

\textbf{FRGC2} \quad The FRGC2 dataset \cite{phillips2009frvt} are divided into train partition and 
validation partition and has 50,000 images in all. The validation partition 
consists of datas from 4,003 subject sessions. A subject session is the set of all images of a person 
including four controlled still images, two uncontrolled still images, and one three-dimensional image 
which contains both a range and a texture image.\par

\section{Future Prospects}

At present, there are two main development directions of single image 3D face reconstruction. 
One is multitasking 3D face reconstruction, which means complete other face related tasks 
and 3D face reconstruction in the meantime, such as face recognition and face alignment.  
So that it can be mutual beneficial to improve the performance of multiple task at the same time. 
The other one is to reconstruct a more refined 3D face that contains vivid and true expressions and details like wrinkles.

\bibliography{Literature_Lib}
\bibliographystyle{plain}

\end{document}